\documentclass{article}

\usepackage{arxiv}

\usepackage[utf8]{inputenc} % allow utf-8 input
\usepackage[T1]{fontenc}    % use 8-bit T1 fonts
\usepackage{hyperref}       % hyperlinks
\usepackage{url}            % simple URL typesetting
\usepackage{booktabs}       % professional-quality tables
\usepackage{amsfonts}       % blackboard math symbols
\usepackage{nicefrac}       % compact symbols for 1/2, etc.
\usepackage{microtype}      % microtypograph
\usepackage{lipsum}
\usepackage{graphicx}
\graphicspath{ {./images/} }

\title{Emotion estimation from video footage with LSTM}

\author{
 Samer Attrah\thanks{Work done while being a student at HAN} \\
  Engineering and Automotive Academy \\
  Hogeschool Van Arnhem en Nijmegen \\
  \texttt{samiratra95@gmail.com} \\
}

\begin{document}
\maketitle
\begin{abstract}
 Emotion estimation in general is a field that has been studied for a long time, and several approaches exist using machine learning. in this paper, we present an LSTM model, that processes the blendshapes produced by the library MediaPipe, for a face detected in a live stream of a camera, to estimate the main emotion from the facial expressions, this model is trained on the FER2013 dataset and delivers a result of 71\% accuracy and 62\% f1-score which meets the accuracy benchmark of the FER2013 dataset, with significantly reduced computation costs. \url{https://github.com/Samir-atra/Emotion_estimation_from_video_footage_with_LSTM_ML_algorithm}
 \end{abstract}


\section{Introduction}

In social robotics, many parts of the robot system get integrated into a single system, usually the more complex the system is, the better it will be at performing its tasks, which are more commonly focused on elderly care or patient care. building the subsystems of the robot in addition to the skeletal and motion structure, are the interaction subsystems and senses such as the speech, vision, and hearing functions, and since human emotion is a big part of any interaction between two or more humans, then it is important for the robot to understand the human emotion, and it happens through many senses such as understanding and comprehending speech, voice tone, facial expressions, body pose, and hand gesture when exists, and some times the gaze direction and face orientation could help in estimating the emotion of a person. In this work, we focus solely on emotion estimation from facial expressions.

Emotion estimation from facial expressions, more commonly referred to as Facial Emotion Recognition (FER), models could be built in three steps, that are 1) face detection: in a camera stream or a photo, and localizing the boundaries of the face, 2) landmark detection: extracting features from the face previously detected, by setting key points as a vector for the most relevant parts of the face, such as the tip of the nose and the ends of the mouth in two dimensional or three-dimensional coordinates, 3) expression classification: which comes after extracting the features by inputting them into a classification model, to predict the emotion represented by the expression shown on the face.

The system we are presenting for emotion estimation is to be implemented as a feedback system, to help drive a conversation or an interaction between a human and a robot, where it enables the robot to detect in real-time the reaction of the human through their facial expression, and based on that could change the topic or the approach of a conversation, or possibly an action it is taking.

The scope of this research was limited by the availability of data and computation resources, and to work with that the task of emotion classification was narrowed to three classes, instead of the possibility of classifying eight of them\cite{ekman2003unmasking}{}, and those three classes with labels happy, sad, for corresponding emotions and label unknown for every other emotion. 

To start building the system will be building the full data processing pipeline, beginning from loading, cleaning, augmenting, feature extracting, and visualizing the data and then it gets inputted into the model for training. and for the inference use of the model will integrate the trained model to the Gaze Project\footnote[1]{\url{https://gitlab.com/Hoog-V/gaze/-/tree/main/gaze_estimator_python_rpi?ref_type=heads}} which will do the face detection/localization and feature extraction parts, then the trained model will be classifying the face expression depending on the features.

Some considerations to be noted are, that the dataset used FER2013\cite{goodfellow2013challenges}{}, is a challenge-dataset i.e. it is not built and optimized for research so it includes a group of non-relevant images such as animations and covered faces which does not give useful information to the model to learn from. also, work will be done on a laptop Graphics NVIDIA GeForce RTX 3050 Mobile. to train and test the model on.

This research is a proof of concept (POC), for the emotion estimation for embedded systems in social robotics full project, the main contributions this research is making are:

\begin{enumerate}
    \item Building a small and cost-efficient model while taking into consideration the spatial and temporal aspects of the facial expression.

    \item Using the MediaPipe library to localize the face and exact features for it, and using the Blendshapes\cite{lugaresi2019mediapipe}{} as features which is rarely the case when it comes to FER applications.
\end{enumerate}

The rest of the paper is organized as follows: section\ref{sec:relatedwork} Related Work, section\ref{sec:methods} Methods, section\ref{sec:ablationstudies} Ablation Studies, section\ref{sec:results} Results, and section\ref{sec:conclusion} Conclusion.

\section{Related Work}
\label{sec:relatedwork}

Facial emotion recognition is an application that has been developed and researched, for a long time\cite{ekman2003unmasking}{} and many methods for face detection and tracking such as the Haar cascade algorithm, and another for feature extraction were proposed such as using the classical algorithms HOG, SIFT\cite{gautam2023facial}{}\cite{kumar2016real}{}, or more recently using the CNN\cite{li2020deep}{} and transformer-encoders to detect the face and extract the features, and the classification models also using CNNs, RNNs or a combination\cite{li2020deep}{}\cite{leong2023facial}{}.

\subsection{Feature extraction}

Feature extraction can be considered as, a special kind of data dimensionality reduction, of which the goal is to find a subset of informative variables based on image data\cite{egmont2002image}{}. the most common approach for extracting quality features from images of faces happens by estimating landmarks on the face, one approach to do that is finding the tip of the nose, and then segmenting it, by
cropping the sphere centred at this tip, and then finding the eyebrows, mouth corners, eye corners and possibly many others on the segmented sphere, and by measuring the distance between the landmarks and changes in their positions, could understand the expression being shown on the face\cite{farkhod2022development}{}.

\subsubsection{History}
Looking back on the evolution of FER, one of the methods used to extract features is the Histogram of Oriented Gradients (HOG)\cite{gautam2023facial}{}, and it works by determining whether each pixel is an edge or not, the edge direction, and the magnitude of the edge. another method is the Scale-Invariant Feature Transform (SIFT)\cite{kumar2016real}{}, which works in four parts these are: 1) construction of the scale space, 2) crucial point localization, 3) orientation assignment, and 4) assigning uniquely maintained fingerprint.

More recently, feature extraction is done using Deep neural networks\cite{li2020deep}{}\cite{leong2023facial}{}, such as fully connected networks (FCN) or convolutional neural networks (CNN), and that is by building a model that consists of a number of layers and hyperparameters, determined by the complexity of the data, and the performance quality required, besides the computation resources the model meant to be implemented on.

\subsubsection{MediaPipe}

In this work, we used the MediaPipe\cite{lugaresi2019mediapipe}{}\cite{lugaresi2019mediapipe2}{} library for facial recognition and tracking, and for feature extraction, in other works such as in\cite{BISOGNI2023104724}{} which suggested two approaches for feature extraction, in one of them, used MediaPipe to extract features from the images, and found that using this library is better than using Deep learning networks such as CNNs, for feature extraction of all 8 emotions exists in the dataset used in that work, in the case of slight face expressions, and the more intense the face expressions get, the smaller the performance gap between the two approaches, until in the most intense case CNNs show better results in the majority of the emotions. 

\begin{figure}
    \centering
    \includegraphics[width=0.25\linewidth]{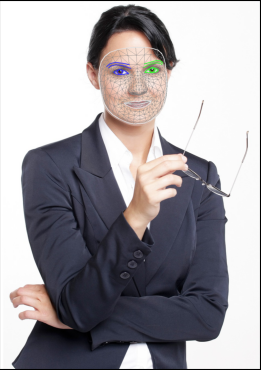}
    \caption{Image annotated with MediaPipe}
    \label{fig:annotatedimage}
\end{figure}

Another work is\cite{savin2021comparison}{}, where it compares MediaPipe to OpenFace, so when it comes to the landmark detection problem OpenFace\cite{7477553}{} uses OpenCV and detects 68 facial landmarks, while MediaPipe uses TensorFlow\cite{abadi2016tensorflowlargescalemachinelearning}{}, to detect 468 landmarks which are arranged in fixed quads and represented by their coordinates (x,y,z). Figure\ref{fig:annotatedimage} shows an image annotated with the landmarks used to detect the expressions on it.

\subsubsection{Blendshapes}

The features to be used, are the second output from the MediaPipe Face landmark detection task model, which is the blendshapes of the frame, which are an approximate semantic parametrization and a simple linear model of facial expression\cite{lewis2014practice}{}, that was first originated in the industry before academic research and was prevalent in computer graphics. Although the blendshape technique is conceptually simple,
developing a blendshape face model is a large and labor
intensive effort, To express a complete range of
realistic expressions, so one face might require more than 600 blendshape to be sufficiently expressed.

A single blendshape construction, was originally guided by the facial action coding system (FACS)\cite{ekman1997face}{}, where using that system can manually code all facial displays, which are referred to as action units and more than 7000 combinations have been observed. FACS action units are the smallest visibly discriminable changes in facial display, and combinations of FACS action units can be used to describe emotion expressions\cite{ekman1993facial}{} and global distinctions between positive and negative expressions.

\begin{figure}
    \centering
    \includegraphics[width=0.75\linewidth]{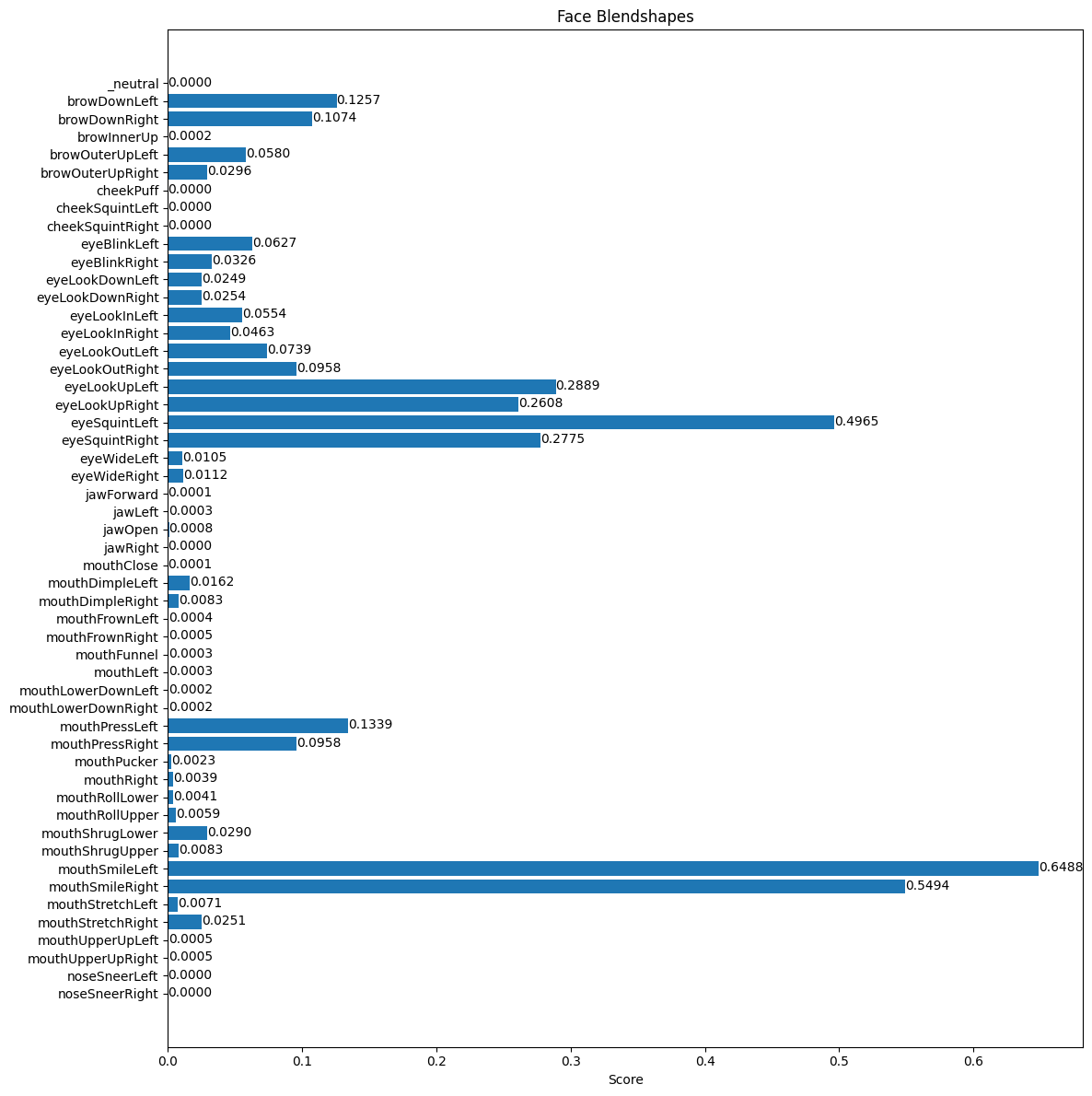}
    \caption{Histogram for the blendshpes of the image in figure1 by MediaPipe}
    \label{fig:blendshapeshistogram}
\end{figure}

MediaPipe uses the ARKit face blendshapes\footnote{\url{https://arkit-face-blendshapes.com/}},
which consists of 52 blendshapes that describe the face parts and expressions with probability scores in the range (0-1), showing the existence of the specific blendshape, shown in Figure\ref{fig:blendshapeshistogram} where it shows the blendshapes histogram for the image in Figure\ref{fig:annotatedimage}

\subsection{Emotion classification}
For emotion classification, there are many approaches that has been used in research, such as Support Vector Machines (SVM)\cite{kumar2016real}{} where it was implemented with linear as well as radial basis function in addition to stochastic gradient descent (SGD) classifier, which obtained results around 95\% accuracy on the Radboud faces Database (RaFD)\cite{langner2010presentation}{}, and in\cite{BISOGNI2023104724}{} presented a comparison between two approaches, that are a MediaPipe-SVM and a CNN-LSTM where the MediaPipe and the CNN are for feature extraction and the SVM and LSTM\cite{computation2016long}{} are for classification, and by experiment showed that the quality of classification based on the quality of features, and since the features extracted by MediaPipe are superior to the ones extracted by CNN, then the classification from the first combination comes with higher quality than the second. 

Another image classification is by using Convolutional Neural Networks (CNN), where in\cite{gautam2023facial}{} they used a sequential model of three convolutional layers, and a dense layer to classify the features inputted, and the method shows high accuracy on the dataset used. 

Since the model being developed is meant to process videos which is a type of sequential data, then as stated in\cite{goodfellow2016deep}{} much as a convolutional network is a neural network that is specialized for processing a grid of values such as an image, a recurrent neural network is a neural network that is specialized for processing a sequence of values or sequential data. so searching for the best type of RNN is a necessary step.

When working with sequential data such as videos, and needs to consider the time dependency between the images, using a recurrent neural network is the best choice for that as in\cite{li2020deep}{}, and using an LSTM which is a special type of RNN comes to address the vanishing gradient and the exploding gradient problems that are common in RNN training, in\cite{jain2018hybrid}{} proved by the comparison of the results, that the network including a CNN part and a RNN part delivers better accuracy, than a network with CNN layers only and that is by around 20\% higher accuracy.

Another work discussing recurrent neural networks (RNN) is\cite{leong2023facial}{}, where it suggested a CNN-LSTM architecture as a facial recognition system that could understand the spatio-temporal properties in a video.

\section{Methods}
\label{sec:methods}

In this section we will discuss our approach to building the emotion estimation system, starting with the dataset used, and the data processing techniques then building the model and optimizing it, and ending with evaluating and testing the performance of the model.

\subsection{Dataset}

Since the model being built is to be used in inference from video, then looking for a video dataset of different facial expressions was the first choice, and from\cite{li2020deep} can find a group of datasets some of them are for videos such as MMI\cite{pantic2005web}{}\cite{valstar2010induced}{} and AFEW7.0\cite{dhall2017individual}{}, but given the limited computation resources that are necessary for the data processing and other constraints, did not include that in the research scope, and decided to use an images dataset.

A few of the choices that were considered are MultiPIE\cite{gross2010multi}{}, and that was because of the quality of its images due to the approach it was made by, and the suitable size of it, also considered the Radboud Faces Database RaFD\cite{langner2010presentation}{} for the structure of the database and the quality of the distribution, another choice was AffectNet\cite{mollahosseini2017affectnet}{} once again that was because of the suitable size and the popularity of it since it is used as a benchmark to evaluate the performance of the models in many research and experiments.

For this work we used the FER2013 dataset\cite{goodfellow2013challenges}{} that was for its simplicity, having 48x48 greyscale images and having a useful number of images and easy to obtain being an open access dataset. 

The dataset was downloaded from Kaggle\footnote{\url{https://www.kaggle.com/c/challenges-in-representation-learning-facial-expression-recognition-challenge/data}} and the data was in a .csv file representation that includes the class, image pixels array grey-scale values, and the data split, as columns and the training examples as rows.

The FER2013 dataset is structured into 7 classes that are happy, sad, angry, afraid, surprise, disgust, and neutral, and the full amount of images are split into three groups training, public test, and private test, which were used as training set, validation set and test set, respectively, and the counts of the images are as shown in Table\ref{tab:datasetsplits}:

\begin{table}[h!]
    \centering
    \begin{tabular}{ c c c c c }
        \hline
        Class - Split & Full & Training & Public test & Private test\\
        \hline
        Happy & 8989 & 7215 & 895 & 879\\
        Sad & 6077 & 4830 & 653 & 594\\
        Angry & 4953 & 3995 & 467 & 491\\
        Afraid & 5121 & 4097 & 496 & 528\\
        Surprise & 4002 & 3171 & 415 & 416\\
        Disgust & 547 & 436 & 56 & 55\\
        Neutral & 6198 & 4965 & 607 & 626\\
        \hline
    \end{tabular}
    \caption{FER2013 dataset: classes and splits}
    \label{tab:datasetsplits}
\end{table}

From the table can notice that 1) The happy, sad, and neutral classes are the highest in count, while the disgust class is so rare. 2) The public test and the private test splits are fit in count to be used as the validation set and test set respectively. in the next subsections will discuss processing techniques on the dataset that will change and restructure the database in a form that best serves the model being built.

\subsection{Data processing and cleaning}

Using the FER2013 dataset for research while it is built for a challenge and a benchmark, comes with some requirements of data processing to make the data fit for producing an emotion estimator for a robotics application, and to do that a few steps of cleaning and processing are required.

\subsubsection{Creating training data classes}

The first step of data processing is to create three splits of the dataset: the training, validation, and test sets, because of the classes' count imbalance, decided not to use the full amount of available training classes' images, but include parts of them and as shown in Table\ref{tab:classcount}:

\begin{table}[h!]
    \centering
    \begin{tabular}{cccccccc}
        \hline
        Class & Happy & Sad & Angry & Afraid & Surprise & Disgust & Neutral\\
        \hline
        Count & 4000 & 4000 & 1500 & 1500 & 1500 & Maximum possible & 1500\\
        \hline
    \end{tabular}
    \caption{Training set classes counts}
    \label{tab:classcount}
\end{table}

Choosing these numbers came because the model being built works for happy, sad, and unknown classes only, instead of the full 8 emotions' classes, so having 4000 for the main emotions will be in favour of balancing the dataset, and choosing the 1500 for most of the others because the sum of them will be 6000 in addition to the disgust class, which amounts to around 400 images, will make a reasonable count of training examples to make the unknown class to the side of the happy and sad.

Looking at Table\ref{tab:datasetsplits} could notice that, including bigger than these numbers of training examples from each class is possible for some of them, but including all could mean having an imbalanced data representation and a less quality distribution.

\subsubsection{Readability by MediaPipe}

To clean the dataset and avoid runtime errors, caused by having an image undetectable by MediaPipe, created a detection program and ran all the images of the training set through mediapipe, and excluded the undetectable ones from the dataset.

This process resulted in diagnosing the following numbers of each class as in Table\ref{tab:undetectablecount}.

\begin{table}[h!]
    \centering
    \begin{tabular}{cccccccc}
        \hline
         Class & Happy & Sad & Angry & Afraid & Surprise & Disgust & Neutral \\
         \hline
         Counts & 377 & 835 & 597 & 616 & 239 & 72 & 261 \\
         \hline
    \end{tabular}
    \caption{MediaPipe undetectable images counts}
    \label{tab:undetectablecount}
\end{table}

\begin{figure}[hbt!]
    \centering
    \includegraphics[width=0.4\linewidth]{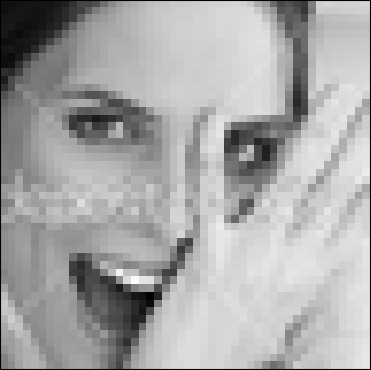}
    \includegraphics[width=0.4\linewidth]{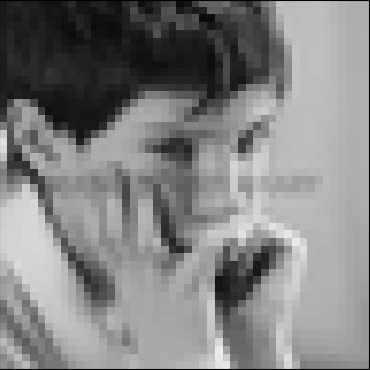}
    \caption{Images undetectable by MediaPipe sample}
    \label{fig:undetectables}
\end{figure}

As described, MediaPipe detects the main parts of the face and estimates the landmark vector for each image, having a part of the face covered will make this process impossible so images such as the ones in Figure\ref{fig:undetectables} are examples for undetectable images.

\subsubsection{Indexing the test set}

This step came after a few experiments resulted in finding an unexplainable high error rate in the model performance, and as a tracking method added a column to the test set which is represented in a .csv file, that assigns a number for each image in the FER2013 dataset, and that number stays with the image through all the steps of data processing, and get transferred to the blendshapes of the image when the blendshapes dataset get created.

This step helps to track the image with a high error, when the model gets evaluated on the blendshapes of the test set, and be able to visualize the image or create different types of plots to check for patterns in the images causing the error.

\subsubsection{Augmenting the training set}

To increase the number of training images and the variety of poses and as a result, improve the generalization of the model decided to augment the data by adding new examples\cite{more2016survey}{} with a random transformation applied to it, the augmentation techniques that were used are:

\begin{enumerate}

    \item Random Horizontal Flip.

    \item Random Rotation by 0.2 x 180 degrees in the clockwise direction, and counter-clockwise direction.

\end{enumerate}

Choosing these techniques comes based on the use case of the model, and since it is being built for a social robot, having a flip will enable the robot to understand reflections, and provide a valid face image but with a different camera angle. while augmenting with the Random rotation would provide robustness to the model predictability, and this is because a tilted or rotated head is a common position for a human, so for the robot to be able to detect that face and classify the expressions shown on it is so helpful and important in real life.

Running the images of the dataset through the augmentation process results in a new extended dataset with more than 20000 images from the three classes described earlier.

\subsubsection{Blendshapes dataset}

Using MediaPipe to detect faces in a video stream, takes many preparations and processing steps for the data before it gets inputted into the classification model, and to build a model that is able to be connected to a face detection program, it needs to be trained on receiving the same output from that program, and for that processing the full dataset into a blendshapes dataset and use it to train the model.

\subsection{Model architecture and training}

The classification model was built fully from the LSTM layer\cite{computation2016long}{} except one last Dense layer with Softmax activation to give a classification score from the last layer output.

The decision to use the LSTM layer to build the model was made for the following reasons:

\begin{enumerate}

    \item The model being built is to be integrated to a video stream and
    classify the faces in it, which means it is a time-series data application and that means the output generated for each input to the model has a dependency on the time of its occurence, so using one type of recurrent neural network (RNN)\cite{goodfellow2016deep}{} is intuitive. Because it considers the past states besides the current input.
    
    \item Another reason for this choice is that LSTM is a type of gated recurrent unit (GRU)\cite{chung2014empirical}{}, which means one of its advantages is having an inside gate that keeps out the non-relevant features from changing the loss in the model and only consider the critical ones.
    
    \item LSTM stands for Long-Short Term Memory, which reflects one advantage for this type of network on the other types of recurrent neural networks, which is the ability for it to keep track for long-term dependencies well, and not only short-term dependencies, as the case with RNN\cite{medsker2001recurrent}{} and GRU networks.
    
\end{enumerate}

After choosing the type of the layer i.e. LSTM, to be mainly used, start with building the architecture of the model, and since there are many conditions to be met, that are accuracy, recall, precision, latency, and model size, and after a few experiments to find the best parameters, decided that using Keras tuner\cite{omalley2019kerastuner}{} as an architecture search framework to find the best values for the hyperparameters, would be more efficient. and setting the search space came based on the experiments and intuition\footnote{\url{https://medium.com/p/1833f774051f}}.

\begin{figure}[hbt!]
    \centering
    \includegraphics[width=1\linewidth]{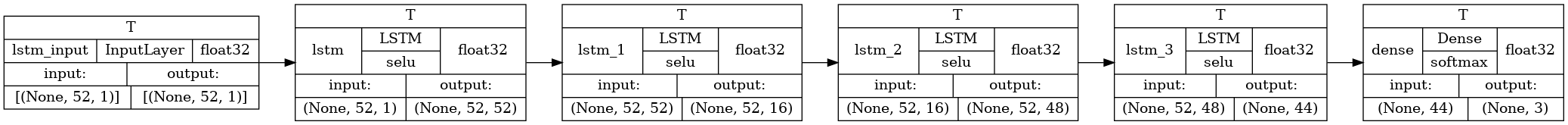}
    \caption{Model structure}
    \label{fig:modelstructure}
\end{figure}

The final model was trained for 5000 epochs which took around two days to finish, using the batch size of 128 and the architecture was as in Figure\ref{fig:modelstructure}, and the optimizer used was AdamW\cite{kingma2014adam}{}\cite{loshchilov2017decoupled}{} with learning rate of 1.09e-06, global clipnorm of 1, and amsgrad was set. callbacks used were checkpoint callback and early-stopping callback.

For the loss function the categorical crossentropy was used, and for that encoded the labels of the images as one-hot vectors, and the metrics for evaluation were the loss, categorical crossentropy, categorical accuracy, and F1-score.

\subsection{Model evaluation}

The process of model evaluation took place on every set of weights get produced by the checkpoint callback, and shows improved training and validation performance metrics. where around every 100 epochs a model weights set gets saved. 

Each model gets evaluated by predicting the labels of the test dataset images and comparing them to the ground truth labels. and a loss, categorical crossentropy,  categorical accuracy, and F1-score get calculated for the model.

\section{Ablation Studies}
\label{sec:ablationstudies}

To have a better understanding of the model and the possibilities for the application being developed, we conducted a few ablation studies, to test for the best number of blendshapes to be used, the type of the model, and the loss function.

\subsection{Choose relevant blendshapes}

To improve the efficiency of the model, lower the computation requirement for it, and be able to deliver quality classification results with less processing time and memory usage, decided to disregard a set of the blendshapes returned from MediaPipe for the face detected in each frame of the video that will be processed by the model, which is 52 according to the ARKit face blendshapes.

This idea came after noticing and as might be obvious in Figure\ref{fig:blendshapeshistogram} that some of the blendshapes do not have a score at all, so tested for the whole dataset and found that some of the blendshapes do not have a score for all the images in the dataset.

From that notice we had two choices to use 1) count the times a certain blendshape have a zero score and disregard it, if that was a high count. 2) set a threshold number and count how many times each of the blendshapes will have a score higher than this threshold.

After testing a few values for the first choice high counts and for the second choice thresholds, decided to take the second choice and use 0.4 as a threshold and set 100 as a high count number, so for every blendshape score if it does not go higher than 0.4 for more than 100 images in the dataset, it gets disregarded.

That choice was made for the following reasons:

\begin{itemize}

    \item Counting the zeros and making a decision based on this criteria, would have resulted a small and non-relevant number of blendshapes being disregarded, and that would not have improved the efficiency of the model after all.

    \item Taking the second choice and setting a threshold higher than 0.4, would have resulted in disregarding blendshapes that have a wide range of probability, which gives quality information to the model being trained to learn from, and that will get the model smaller but cut from the accuracy and performance.

    \item Using a higher count than 100, would have resulted in a significant cut down in the numbers of the blendshapes being used, and that would make it more difficult for the model to learn.

\end{itemize}

These reasons listed were also proven by experimenting i.e. training models on a different data collection, and testing its performance and size, then making the decision for every number mentioned.

That process resulted in using a count of 27 blendshapes out of the 52, which will decrease the size of the model while keeping the performance metrics such as the accuracy and f1-score the same.

\begin{figure}[hbt!]
    \centering
    \includegraphics[width=0.75\linewidth]{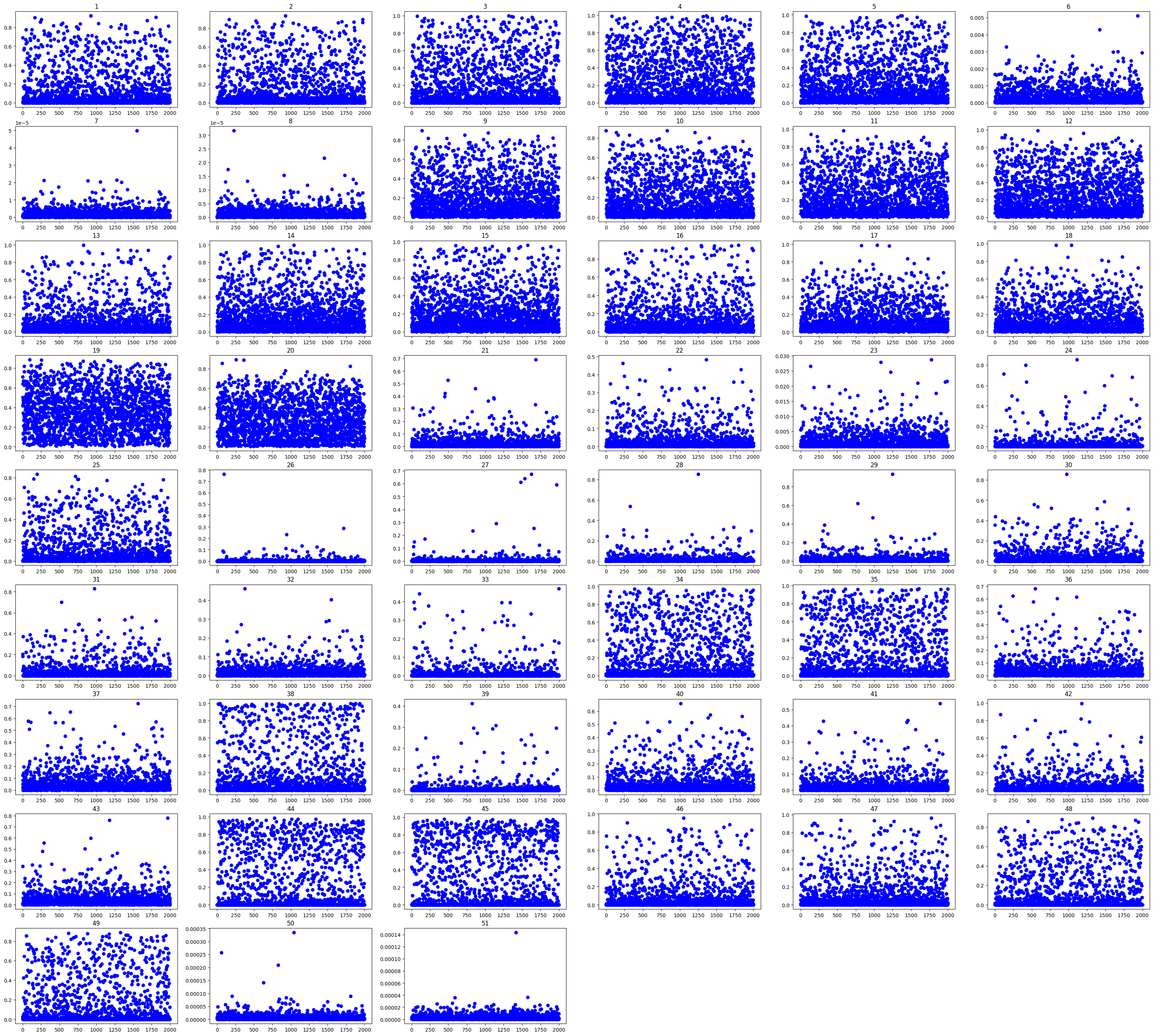}
    \caption{Blendshapes scatter plot for a sample image}
    \label{fig:blendshapesscatter}
\end{figure}

For further clarification, Figure\ref{fig:blendshapesscatter} shows the scatter plot for the 52 blendshapes and how many of them do not have high values which result in low information that could be neglected.

\subsection{Dense neural network model}
To get a faster model and shorter prediction time (latency) than of the LSTM network, decided to build another network that is less complex and for that chose the fully connected (Dense) layer as the only layer in the whole model. after a few optimization experiments, the metrics showed high performance from the network, but when it was integrated into the system and received the camera stream, it did not give acceptable stability in the results and kept oscillating between the classes without changing the view in the camera, besides not giving sufficiently accurate predictions.

\subsection{CCE and MSE loss function}

In the earlier stages of the research used the Mean Squared error\cite{ref1}{} loss function for training the LSTM model, but because the MSE function works by calculating the distance between the prediction and the ground truth, instead of calculating the probability difference, which is fit for regression applications, had to change the loss function after a certain point.

\begin{equation}
    MSE = \frac{1}{n} \sum_{i=1}^{n} (y_i - \hat{y}_i)^2
    \label{eq:mse}
\end{equation}

In comparison to the cross-entropy, which finds the difference between the probability of having the correct category assigned to the face, and the ground truth label. which fits for classification applications. and since the application includes three classes then using the Categorical Cross-Entropy was the choice.

\begin{equation}
    CCE = -\sum_{i=1}^{N} y_i \log(\hat{y}_i)
    \label{eq:categorical_cross_entropy}
\end{equation}

for possible further improvement, suggest using the Categorical Focal Cross-Entropy\cite{lin2017focal}{}, which will take into consideration the class imbalance in the dataset. although it was originally built for object detection, experiments showed that it also improves the classification models' results.

\begin{equation}
    FL = - \alpha (1-\hat{y}) ^\gamma CCE(y, \hat{y})
    \label{eq: Categorical focal cross-entropy}
\end{equation}

\section{Results}
\label{sec:results}

The results that were obtained from the model were close to the classification benchmark of the dataset\footnote{\url {https://paperswithcode.com/sota/facial-expression-recognition-on-fer2013}} with no extra training data and using a non-transformer neural network.
The best results obtained from the model on the test set were as follows:

\begin{itemize}

    \item Loss = 0.6238
    \item Categorical crossentropy = 0.6235
    \item Categorical accuracy = 0.7199
    \item F1-score = 0.6298

\end{itemize}

and the confusion matrix for the model as in Table\ref{tab:confusionmatrix} shows the classes as 0:happy, 1:unknown, and 2:sad.

\begin{table}[h!]
    \centering
    \begin{tabular}{c|cccc}
        \hline
         & Predicted \\
         \hline
        Ground truth & Class & 0 & 1 & 2 \\
          & 0 & 251 & 35 & 5 \\
          & 1 & 110 & 850 & 150 \\
          & 2 & 21 & 140 & 84 \\
    \end{tabular}
    \caption{Confusion matrix}
    \label{tab:confusionmatrix}
\end{table}

Experimenting with different architectures, hyperparameters and training for a longer time
did not give improved metrics, but the confusion matrix keeps
oscillating between an improvement in the happiness class classification
one time, and for the sadness class classification another and the one above is biased to the happiness class. and for that considered assigning class weights for each one, to direct the model focus to the happy and sad classes to pay them more attention and give less weight to the unknown class which is easily classified in all experiments. but due to some limitations did not experiment with class weights.

When the model gets integrated to the system camera stream like in the demo video\footnote{\url{https://youtu.be/RdcjA9ScBmI}} it indeed gives
a stable output for happiness and sadness passing by the unknown class
when the face look changes, and for all other facial poses the unknown
class is the output.

\section{Conclusion}
\label{sec:conclusion}

We worked on building a model to detect faces in a camera video stream, then extract blendshapes as features from the face, and classify the expressions on the face to the emotion that it represents. and proved that a 4-layer LSTM model is a suitable architecture to classify the blendshapes of the faces in the frames of the camera stream, showing good stability with respect to time. and note that it shows good results in the video stream even after the model gets trained on an image dataset. besides the model delivers results equal to the benchmark of the dataset with no loss in accuracy through the pipeline of feature extraction and classification, which saves a lot of memory and computation in comparison to other available methods.

For future work, will be using an improved dataset and computation and work on delivering a smaller model in size and higher in performance.

\section*{Acknowledgments}

Thanks to professor Marijn Jongerden, Jeroen Veen and Dixon Devasia for their help and guidance along the way, and thanks to Victor Hogeweij for his contribution to the Gaze project. Also, thanks Bhupinder Kaur, An Le, and Muhammad Reza for their insight and support at the early stages of the work.

\bibliographystyle{unsrt} 
\bibliography{latexbib}

\end{document}